\documentclass{article}

\usepackage{arxiv}

\usepackage[utf8]{inputenc} 
\usepackage[T1]{fontenc}    
\usepackage{hyperref}       
\usepackage{url}            
\usepackage{booktabs}       
\usepackage{amsfonts}       
\usepackage{nicefrac}       
\usepackage{microtype}      
\usepackage{graphicx}
\usepackage{natbib}
\usepackage{doi}

\title{Learning from Waste: Machine Learning for Health Risk Prediction and Computer Vision--Based Sorting in Ghana}

\date{August 4, 2026}

\author{
\href{https://orcid.org/0009-0005-4873-7719}{\includegraphics[scale=0.06]{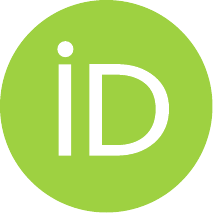}\hspace{1mm} Hilda Adwubi Osei} \\
Department of Industrial Engineering\\
Kwame Nkrumah University of Science and Technology\\
Kumasi \\
\texttt{hildaosei109@gmail.com}
\And
\href{https://orcid.org/0009-0005-4030-517X}{\includegraphics[scale=0.06]{orcid.pdf}\hspace{1mm} Catherine Tenewaa Osei} \\
Department of Nursing\\
Kwame Nkrumah University of Science and Technology\\
Kumasi \\
\texttt{oseicathryn3@gmail.com}
\And
\href{https://orcid.org/0009-0004-3720-3642}{\includegraphics[scale=0.06]{orcid.pdf}\hspace{1mm} Desdemona Yaa Asobayire} \\
Department of Mechanical Engineering\\
University of Nottingham\\
NG7 2RD, United Kingdom \\
\texttt{desdemonaasobayire952@gmail.com}
}

\renewcommand{\shorttitle}{Learning from Waste}

\hypersetup{
pdftitle={Improper Waste Management and Data-Driven Approaches to Segregation in Atonsu, Kumasi},
pdfsubject={cs.LG, cs.CY},
pdfauthor={Hilda Adwubi Osei},
pdfkeywords={waste management, machine learning, public health, Ghana,  image classification,  predictive modeling, community-based research},
}

\usepackage{float}
\begin{document}
\maketitle

\begin{abstract}
The inappropriate disposal of solid waste remains a significant public health and environmental concern worldwide, including in Ghana. Poor sanitation and improper municipal waste management practices in Ghana contribute to substantial economic costs and thousands of avoidable deaths annually. In 2022, a field study conducted in Atonsu, Kumasi, Ghana, reported a community-perceived relationship between household waste disposal practices and particular patterns of illness. However, this association was identified only descriptively, based on interview responses and questionnaire percentages, without quantitative modeling or validation.

This study extends that earlier investigation through the application of two data-driven approaches. First, a Random Forest classifier was developed to predict illness categories using household waste disposal practices alongside demographic survey information. On a held-out group of respondents who reported illness ($N=69$), the model obtained a macro F1 score of 0.63, with waste disposal method emerging as the most important substantive predictor of illness type. Second, a MobileNetV2-based image classification model was designed to enable automated waste sorting through visual recognition. The model achieved 88.2\% accuracy and a macro F1 score of 0.87 on the test set ($N=415$).

The vision-based approach offers a relatively affordable, camera-driven alternative to the multi-sensor mechanical sorting system proposed in the original work, making it potentially more suitable for deployment in resource-constrained settings. Taken together, the findings provide quantitative evidence for a community health relationship that had previously been documented only through qualitative and descriptive analysis. They also demonstrate the potential for practical automated waste-sorting technologies in environments with limited resources. Importantly, the results further illustrate that technological performance, by itself, does not guarantee meaningful improvements in public health or environmental outcomes; effective institutional support and implementation are equally necessary.
\end{abstract}

\keywords{waste management \and machine learning \and public health \and Ghana \and image classification}

\section{Introduction}
\label{sec:introduction}
The amount of municipal solid waste generated globally continues to grow rapidly, driven largely by urban expansion, population growth, and shifts in consumption behavior. The World Bank's "What a Waste 3.0" report estimates that global municipal solid waste generation totaled approximately 2.56 billion tonnes in 2022. Without significant changes to current trends, this volume is expected to increase substantially, reaching an estimated 3.86 billion tonnes by 2050 \citep{worldbank2026wwaste}.

A disproportionate share of this burden falls on low and middle income countries, where waste management infrastructure has not kept pace with the rate of waste generation, and where a substantial fraction of waste is disposed of through open dumping, informal burning, or unmanaged landfilling rather than through regulated collection and treatment systems, approximately 30\% of global waste was mismanaged in this way in 2022, a share projected to fall to around 20\% by 2050 \citep{worldbank2026wwaste}. These disposal practices carry direct consequences for public health and environmental quality, particularly, in rapidly urbanizing communities where waste management systems are still developing.

In Ghana, poor sanitation and waste management already impose substantial economic and public health costs. A 2012 World Bank Water and Sanitation Program (WSP) study estimated that poor sanitation costs Ghana's economy approximately GH¢420 million annually, equivalent to 1.6 percent of national GDP, with 74 percent of this cost attributable to premature mortality \citep{wsp2012ghana}. The same study attributed an estimated 19,000 annual deaths in Ghana to diarrheal disease, of which nearly 90\% is directly attributable to poor water, sanitation, and hygiene (WASH) conditions \citep{wsp2012ghana}. These national-level figures reflect a broader pattern in which improper waste disposal, including open burning, indiscriminate dumping, and disposal into water bodies, contributes directly to disease transmission and environmental degradation in urban and peri-urban Ghanaian communities.

These national patterns are visible at the community level in Atonsu, Kumasi, where the present study was conducted. An initial field investigation conducted in 2022 identified improper waste disposal as a driver of recurrent illness in the community. This was based on the observation that frequent hospital visits by residents corresponded with specific, repeated diagnoses, including cholera, malaria, typhoid, diarrhea, and colds and flu, many of which were traceable to how residents disposed of their waste. Structured interviews and questionnaires administered to residents of Atonsu Bukuro found that most respondents disposed of waste through open burning, dumping in streams, or discarding in bushes, with only a minority relying on regulated bin collection. Residents linked each disposal method to a distinct pattern of health complaints: waste dumped in bushes was associated with malaria and snake bites; waste dumped in streams was associated with diarrhea, cholera, and dysentery; and smoke from open burning was associated with headaches, nausea, and heart-related issues. This fieldwork established a clear, community-reported link between disposal methods and illness, but did not go further than descriptive and anecdotal analysis, leaving open the question of whether this relationship could be modeled and quantified systematically, and whether interventions such as automated waste sorting could be feasibly implemented in a resource-constrained setting. 

This study extends the original field investigation with two data-driven models aimed at addressing the gaps identified above. First, we developed a Random Forest classifier that predicts the illness category from self-reported disposal methods and demographic variables using survey data collected from residents across Atonsu's eight sub-areas. Evaluated on a subset of respondents who reported illness, the model achieved a macro F1 score of 0.63 across four illness categories — a modest but meaningful result given the small sample size (N = 69) and the inherent noisiness of self-reported health outcomes, and one that identifies disposal method as the dominant predictive feature. Second, instead of the original study's conceptual, multi-sensor mechanical waste-sorting design, we developed a vision-based waste classifier using transfer learning on MobileNetV2, fine-tuned on the TrashNet dataset, achieving 88 percent test accuracy and a macro F1 score of 0.87 across waste categories. Together, these two models demonstrate that (1) the community-reported link between disposal method and illness can be partially recovered through supervised learning on real survey data, and (2) camera-based waste classification offers a simpler, more readily deployable alternative to the multi-sensor sorting mechanism originally proposed, without requiring the combination of infrared, capacitive, inductive, and optical sensors envisioned in the 2022 concept design.

The rest of this study is structured as follows. Section 2 reviews the related work on waste disposal and health outcomes, predictive modeling of environmental survey data, and automated waste sorting approaches. Section 3 describes the study area and the background of the original field investigation conducted in 2022. Section 4 details the methodology for both the survey-based and image classification models. Section 5 presents the results for each model, and Section 6 discusses their combined implications. Section 7 outlines the limitations of this study, and Section 8 concludes with recommendations for future research.

\section{Related Work}
\label{sec:related-work}
 \subsection{Waste Disposal and Health Outcomes in Sub-Saharan Africa}
 Prior research across Sub-Saharan Africa has established consistent links between improper solid waste disposal and community health outcomes, though largely through descriptive and qualitative methods. A recent systematic review of 27 studies spanning eight Sub-Saharan African countries, including Ghana, synthesized this literature into a conceptual framework connecting disposal practices to health outcomes through soil, air, and water contamination pathways \citep{sswaste2026review}. Within Ghana specifically, a focus-group study in Accra found out that residents directly attributed malaria, cholera, and diarrhea to improper waste disposal, alongside respiratory and eye irritation from waste burning \citep{kanhai2019accra}: a pattern closely mirroring the four disposal-linked illness categories identified in the original 2022 field investigation this paper builds on. Cross-sectional household surveys have also been used to document disposal practices and their demographic correlates in Ghanaian communities, including a 700-household study in Sunyani that examined waste handling behavior alongside socioeconomic factors \citep{agyeimensah2017sunyani}, and a semi-quantitative risk assessment in rural Ghana that identified dumpsites and uncontrolled burying as the highest-risk disposal practices for infectious and vector-borne disease \citep{ruralghana2022risk}. Across this body of work, however, the relationship between disposal method and illness is established primarily through description, correlation, or expert-assigned risk scores, rather than through predictive modeling capable of quantifying the strength of these associations from individual-level data: a gap this paper addresses directly.
 
 \subsection{Predictive Modeling on Environmental/Behavioral Survey Data}
 Machine learning approaches have increasingly been applied to survey-based health and environmental data across Sub-Saharan Africa, though typically at large regional scale. A recent study compared five supervised models, Random Forest, Decision Tree, XGBoost, Logistic Regression, and an artificial neural network, to predict household sanitation facility access using Demographic and Health Survey (DHS) data spanning 34 Sub-Saharan African countries and over 500,000 households, using SHAP analysis to interpret feature contributions \citep{dhs2025sanitation}. Broader reviews of DHS-based machine learning applications similarly report Random Forest and Logistic Regression as standard choices for classification tasks on demographic and health survey data in low and middle income country contexts, often paired with resampling techniques such as SMOTE to address class imbalance \citep{dhs2024review}. At smaller scales, methodological work on Random Forest performance in small, imbalanced datasets have shown that careful hyperparameter tuning and class-weighting can meaningfully improve prediction on minority classes even when sample sizes are limited \citep{smallrf2021}, a finding echoed in healthcare applications where Random Forest has outperformed support vector machines and boosting methods on imbalanced multi-class disease prediction tasks \citep{imbalancedrf2011}. The present study differs from this literature primarily in scale and granularity: rather than modeling sanitation access or disease risk across tens of thousands of households at a national or regional level, it applies the same class of models to a single community's field survey, directly linking individual-level disposal method to self-reported illness category. This localized approach trades statistical power, reflected in the modest sample size and macro F1 score reported in Section 5, for a level of contextual specificity that large-scale DHS-style studies, by design, do not capture.
 \subsection{Automated Waste Sorting: Sensor-Based and Vision-Based Approaches}
 Efforts to automate waste sorting have followed two broad technical paths: sensor-based mechanical systems and, more recently, vision-based classification. At the academic and small-scale end, several West African engineering projects have proposed sensor-driven sorters closely resembling the multi-sensor concept originally developed for this study: an Arduino-controlled sorter using infrared and inductive sensors to separate metals from non-metals on a conveyor system \citep{unn2022sorter}, and a related design using inductive and capacitive sensors to segregate metal, glass, and plastic streams \citep{ijraset_sorter}. These designs mirror the sensor combination, infrared detection, inductive metal sensing, capacitive material sensing, and servo-actuated bin selection, proposed in the original 2022 conceptual design this paper builds on. At industrial scale, commercial sensor-based sorting systems extend this same principle using near-infrared spectroscopy, visible-light spectroscopy, and electromagnetic induction to sort mixed waste streams at high throughput \citep{sbs2022industrial}, but require capital and maintenance infrastructure well beyond what a single community-level intervention could sustain.

Vision-based waste classification offers an alternative that trades physical sensor hardware for a camera and a trained model. The TrashNet dataset \citep{yang2016trashnet}, comprising roughly 2,500 labeled images across six waste categories, has become a standard benchmark for this approach, with published results spanning a wide performance range: a ResNet18 architecture with a channel-attention module achieved 95.87 percentage accuracy \citep{zhang2021ctr}, a deeper ResNet variant (RWNet-101) reached 89.9 percentage without specifically addressing the dataset's class imbalance \citep{rwnet2022}, a transfer-learning approach (RecycleNet) achieved 81 percentage on a limited subset of the data \citep{recyclenet2018}, and a convolutional network trained from scratch without transfer learning reached only 62.2 percentage \citep{cnn_baseline}. The MobileNetV2 model developed in this study, evaluated on a proper 70/15/15 train/validation/test split, achieved 88 percentage test accuracy and a macro F1 score of 0.87, comparable to mid-tier published TrashNet results and substantially above from-scratch or non-transfer-learning baselines, while falling short of top-performing attention-augmented architectures optimized specifically for this benchmark.

This literature suggests that vision-based classification can achieve a practically useful accuracy on standard waste-sorting benchmarks without the physical sensor infrastructure required by mechanical sorting designs. This motivates the central methodological shift in this study: replacing the original study's proposed multi-sensor mechanical sorter with a camera-based classifier as a more readily deployable alternative for a resource-constrained community setting.
 \subsection{Municipal Solid Waste Management in Ghana and West Africa}
 The challenges observed in Atonsu reflect broader, well-documented failures in municipal solid waste management across Ghana's major cities. Nationally, Ghana generates an estimated 12,710 tonnes of solid waste daily, of which only about 10 percent is collected and disposed of at designated sites \citep{ghanamsw2021}. In Accra and Kumasi specifically, the country's two largest cities, together generated roughly 4,200 tonnes of waste daily at a collection rate of approximately 70 percent, uncollected waste accumulates in open dumps, choked gutters, and unmanaged dumpsites, particularly in low-income and peri-urban settlements where access to formal collection routes is limited \citep{owusuansah2015conundrum}. Notably, the same source reports Ghana Health Service data attributing approximately 80 percent of cholera and diarrhea cases nationally to communities facing waste management challenges, an independent, national scale corroboration of the disposal-illness relationship this paper's survey model investigates at the community level. A case study of the Kumasi metropolis specifically identified inadequate budgetary allocation and weak enforcement of local sanitation by-laws as central drivers of the city's waste management failures, despite cities spending between 30 percent and 50 percent of their operational budgets on solid waste services \citep{kumasiswm2019, ghanamsw2021}. Studies of local government roles in Kumasi Metropolitan Assembly's waste management strategy similarly point to institutional and enforcement gaps rather than purely technical shortfalls \citep{kma_governance2021}. This body of work establishes the structural, city-level roots of the problem investigated in this paper, but stops short of connecting municipal-level failures to individual household disposal behavior and specific illness outcomes, the gap this study's survey-based predictive model addresses directly at the community scale.
 \subsection{Community-Level and Citizen-Science Environmental Monitoring}
 Beyond formal institutional monitoring, community-level and citizen-science approaches have become an established method for environmental data collection in low and middle income countries, particularly for water and air quality. A review of citizen-science water quality monitoring programs across LMICs found that such programs reliably generate useful data for national-level physicochemical and ecological reporting, though data quality declines for more technical parameters, and only 22 percent of programs that identified pollution went on to report it to authorities or take further action \citep{castro2024cswater}, underscoring that community-generated data does not automatically translate into intervention. Smartphone-based approaches in particular have expanded the feasibility of community-level environmental monitoring, using mobile devices already present in many households as data-collection tools rather than requiring dedicated sensor equipment \citep{smartphone2019review}. This literature is well developed for water and air quality but has rarely been extended to solid waste specifically, and rarely pairs community-level data collection with a trained predictive or classification model of the kind developed in this paper. The vision-based classifier presented here extends this smartphone-enabled monitoring paradigm to waste sorting, while the survey-based model demonstrates that community-collected behavioral data can support predictive analysis beyond simple descriptive reporting, directly addressing the data-to-insight gap this literature identifies as a persistent limitation of community monitoring efforts.

\section{Study Area and Background}
\label{sec:study-area}

\subsection{Atonsu, Kumasi}
This study is set in Atonsu, a community within the Kumasi Metropolitan Assembly, situated northeast of Kaase in Ghana's Ashanti Region. Atonsu sits at an elevation of approximately 457 meters above sea level and comprises eight sub-areas: Atonsu Bukuro, Atonsu S-line, Atonsu Monacco, Atonsu Station, Atonsu Agogo, Atonsu Last Stop, Atonsu New Site, and Atonsu Kuwait. At the time of the original 2022 field investigation this study builds on, Atonsu's population was estimated at 8,790\footnote{This figure reflects a localized estimate available at the time of the original 2022 field investigation; no Atonsu-specific figure appears in Ghana's 2021 Population and Housing Census, which reports population at the metropolitan level.}. Kumasi Metropolitan Assembly as a whole recorded a population exceeding 3.3 million in Ghana's 2021 Population and Housing Census \citep{ghanacensus2021}, reflecting the broader rapid urbanization pressures that shape waste management challenges across the metropolis, as discussed in Section 2.4. Atonsu is a well-established residential community with a mix of retail stores, schools, churches, pharmacy shops, and health care centers, including facilities such as Hart Adventist Hospital and Kumasi South Hospital, both of which served as points of contact during the original field investigation's problem identification process, described in Section 3.2.

\subsection{Problem Identification}
The problem addressed in this study was first identified inductively, through repeated observation rather than a predetermined research question. Frequent hospital visits by residents of Atonsu prompted an investigation into the community's most commonly diagnosed illnesses, revealing a consistent and recurring pattern: cholera, malaria, typhoid, diarrhea, colds and flu, and snake bites appeared disproportionately often among patients. This initial observation raised the hypothesis that these health outcomes were not incidental, but connected to how residents of Atonsu managed and disposed of household waste, a hypothesis that structured the subsequent field investigation conducted in 2022.

To test this hypothesis, data were collected using four complementary methods. Internet-based research provided background information on global and national waste management practices and their documented health consequences. Verbal interviews were conducted directly with residents across Atonsu's sub-areas, allowing for open-ended discussions of disposal habits, perceived risks, and lived health experiences that structured questionnaires alone might not capture. Given the difficulty of reaching every resident in person, online questionnaires were additionally administered to extend coverage across a broader cross-section of the community; a copy of this questionnaire is provided in Appendix A. Reference texts on waste management and recycling \citep{riley2008, rogoff2013solid} supplemented the field data with an established technical context on disposal methods and their environmental implications.

Across approximately 1,000 residents engaged through these combined methods, 80 percent acknowledged that improper waste management was a significant problem within their community, an early indication that the issue was widely felt rather than isolated to a small number of complaints. Residents reported four dominant disposal practices: open waste burning (47\%), dumping in nearby streams or water bodies (21\%), use of bins collected through Zoomlion, Ghana's primary waste-collection service provider (13\%), and other or mixed methods (19\%). These four practices and the health complaints residents associated with each formed the empirical foundation of a four-case disposal illness framework central to this study. Case A (bush dumping) was linked to malaria and snakebites, reported by 59\% of residents who experienced related health problems. Case B (stream dumping) was linked to diarrhea, cholera, and dysentery, reported by 24\%. Case C (open burning) was linked to headaches, nausea, and heart-related complaints, reported by 14\%. Case D (bin use) was linked not to illness but to a financial burden, with 3\% of residents citing the cost of bin collection services as their primary concern, a reminder that even the comparatively safer disposal method carried its own barrier to adoption.

This four-case framework, while grounded in real community-reported data, remained descriptive and correlational in nature: it identified an association between disposal method and health outcome without quantifying the strength, consistency, or predictive value of that relationship at the individual level. This limitation directly motivates the survey-based predictive model developed in Section 5.1, which re-analyzes disposal and demographic data through a supervised classification framework to systematically test whether disposal method can predict illness category beyond what descriptive percentages alone can show.

\section{Methodology}
\label{sec:methodology}

\subsection{Data Collection}
Two distinct datasets underpin the models presented in this paper: an original field survey collected specifically for this study, and a publicly available image dataset used as-is.

Survey data were collected using the KoboToolbox, a mobile data collection platform suitable for field research in resource-constrained settings. Sampling was stratified across Atonsu's eight sub-areas (Bukuro, S-line, Monacco, Station, Agogo, Last Stop, New Site, and Kuwait) to ensure geographic representativeness within the community, and bin users, a minority disposal method group in the raw respondent pool, were deliberately oversampled to avoid underrepresentation in downstream modeling. Following collection, the raw survey data underwent a structured cleaning pipeline: records were schema locked to the set of expected columns, categorical values were standardized in casing, and known label inconsistencies were corrected (for example, "Sline" was mapped to "S-line" and "Illeterate" to "Illiterate"). Duplicate records were removed based on a unique respondent identifier, rows missing the target illness variable were dropped, a small number of remaining categorical missing values were imputed using the modal value for that column, and the dataset was filtered to include only respondents who had provided consent. This pipeline reduced the raw dataset to 470 cleaned respondent records used for all subsequent modeling.

The image classification model uses the TrashNet dataset \citep{yang2016trashnet}, a publicly available collection of approximately 2,500 labeled waste images across six categories. TrashNet was used in its original, unmodified form rather than being supplemented with images collected specifically from Atonsu; this decision and its implications for real-world deployment are discussed further in Section 7 (Limitations)

\subsection{Survey Predictive Model}
\label{sec:survey-model}

The survey predictive model addresses a single core task: predicting a respondent's reported illness category from their disposal method and demographic characteristics. Three candidate classifiers were evaluated: logistic regression, decision tree, and random forest, using 5-fold stratified cross-validation on the training partition of the cleaned dataset ($N=376$, stratified 80/20 split from the full $N=470$ dataset with the remaining $N=94$ held out as an untouched test set), which includes a ``None reported'' category for respondents who did not report illness alongside four illness categories corresponding to the case framework introduced in Section~\ref{sec:study-area}. In their untuned, default parameter form, logistic regression achieved the strongest cross-validated performance (macro F1 = 0.665), followed by random forest (macro F1 = 0.648) and decision tree (macro F1 = 0.556).

All three models were then tuned via a grid search over 5-fold stratified cross-validation. The best configuration of logistic regression (\texttt{C=0.1}, \texttt{class\_weight='balanced'}) achieved a cross-validated macro F1 of 0.678. The best configuration of RF (\texttt{class\_weight=None}, \texttt{max\_depth=5}, \texttt{min\_samples\_leaf=10}, \texttt{n\_estimators=100}) achieved a cross-validated macro F1 of 0.697, a modest edge over logistic regression that is small relative to the fold-to-fold variability observed in the untuned comparison. Random Forest was selected as the final model both for this slight performance edge and, more importantly, for its interpretability. Random Forest's feature importance scores provide a direct, model-native way to identify which predictors drive illness classification, a property central to this study's goal of testing whether the disposal method meaningfully predicts illness outcome. Notably, the grid search did not select class weighting for the final Random Forest model, indicating that constraining tree depth and leaf size (\texttt{max\_depth=5}, \texttt{min\_samples\_leaf=10}) was more effective than reweighting classes for this dataset.

Two distinct evaluation results are reported for this model, corresponding to two different analytical questions that are kept explicitly separate throughout this study. The first is the cross-validated macro F1 score of 0.697, which was computed during model selection on the $N=376$ training partition via 5-fold cross-validation on the full 5-class problem. The second is the held-out test-set performance, which was evaluated once on data never used in training or tuning ($N=94$ overall; $N=69$ when restricted to respondents who reported illness, Cases A through D only, as defined in Section~\ref{sec:study-area}). The full test-set and sick-subset results, including per-class precision, recall, and F1, are reported in Section~\ref{sec:results}. These cross-validation and test-set numbers should not be conflated: the former reflects the performance during model selection, while the latter reflects generalization to genuinely unseen data.

The feature importance analysis of the final Random Forest model identified \texttt{fell\_sick\_Yes} and \texttt{fell\_sick\_No} as the two most important features by a substantial margin, an expected result, since this field near-deterministically separates the ``None reported'' class from all illness categories. Among the remaining features, importance was concentrated in the four disposal-method indicator variables in descending order: \texttt{disposal\_method\_Burn}, \texttt{disposal\_method\_Bush/Open Dump}, \texttt{disposal\_method\_Stream}, and \texttt{disposal\_method\_Bin}, with demographic features such as age, education, occupation, sex, and sub-area of residence contributing comparatively little. This finding is consistent with and provides the first quantitative complement to the descriptive disposal illness associations reported by residents in the original 2022 field investigation (Section~\ref{sec:study-area}).

\subsection{Image Classification Model}
\label{sec:image-model}
The image classification model was developed using transfer learning on MobileNetV2, a convolutional neural network architecture pretrained on ImageNet and widely used for lightweight image classification tasks suited to resource-constrained deployment \citep{sandler2018mobilenetv2}. Development proceeded in two stages. An initial exploratory model was trained with the MobileNetV2 base frozen, using only the pretrained convolutional layers for feature extraction while training a new classification head on top, a standard approach for adapting a pretrained network to a new task with limited data.

During this exploratory stage, a methodological error was identified and corrected: an additional \texttt{Rescaling(1./255)} preprocessing layer was applied to input images in sequence with MobileNetV2's own built-in \texttt{preprocess\_input} function, which independently rescales pixel values to the range expected by the pretrained network. Applying both in sequence compressed pixel values well outside this expected range, degrading the effectiveness of the pretrained features and reducing model performance. This issue was identified through inspection of preprocessing outputs during development and corrected by removing the redundant \texttt{Rescaling} layer, leaving \texttt{preprocess\_input} as the sole normalization step. This finding and its correction are reported here directly, both for transparency and because it illustrates a preprocessing pitfall relevant to other transfer-learning applications using pretrained Keras architectures.

Following this correction, a proper 70/15/15 train/validation/test split was constructed from the TrashNet dataset (Section~\ref{sec:methodology}), and the model was retrained from this clean split rather than reusing data partitions from the earlier exploratory run. Training proceeded in two phases: first, with the MobileNetV2 base frozen, training only the classification head; and then, with a subset of the base layers unfrozen for fine-tuning at a reduced learning rate, allowing the pretrained features to adapt further to the TrashNet domain while limiting the risk of catastrophic forgetting. Model selection and hyperparameter decisions were made using the validation split only; the held-out test split was evaluated exactly once, after training was finalized, to avoid any risk of indirect overfitting to test-set performance through repeated evaluation.

On this held-out test set, the fine-tuned model achieved a test accuracy of 88\% and a macro F1 score of 0.87 across the six TrashNet material categories. As discussed in Section~\ref{sec:related-work}, this result falls within the range of published TrashNet benchmarks, exceeding from-scratch convolutional baselines and comparable transfer-learning approaches, while remaining below the top-performing architectures specifically optimized with attention mechanisms for this benchmark. This model is intended to replace the multi-sensor mechanical waste-sorting concept originally proposed in the 2022 field investigation (Section~\ref{sec:study-area}) with a camera-based alternative requiring substantially less specialized hardware.

\section{Results}
\label{sec:results}

\subsection{Survey Model Results}
\label{sec:survey-results}

On the held-out test set ($N=94$), the final Random Forest model achieved an overall accuracy of 0.72 and a macro F1 score of 0.70 across all five categories, including the near-deterministic ``None reported'' class. The per-class performance on this full test set was as follows: Case A (malaria/snake bites) achieved precision 0.73, recall 0.55, and F1 0.63 ($n=20$); Case B (diarrhoea/cholera/dysentery) achieved precision 0.53, recall 0.53, and F1 0.53 ($n=15$); Case C (headache/nausea/heart disease) achieved precision 0.57, recall 0.74, and F1 0.64 ($n=23$); Case D (financial loss/bin use) achieved precision 0.78, recall 0.64, and F1 0.70 ($n=11$); and the ``None reported'' class achieved perfect precision, recall, and F1 (1.00, $n=25$), reflecting its near-deterministic relationship to the \texttt{fell\_sick} survey field.

Restricting evaluation to the subset of respondents who reported experiencing illness ($N=69$, cases A through D only), the subset most relevant to testing whether disposal method predicts illness \emph{type}, accuracy fell to 0.62 and macro F1 to 0.63, with per-class precision, recall, and F1 identical to the full-set values reported above, since these are the same underlying predictions restricted to the sick-respondent rows. The weighted average F1 across this subset was 0.62.

The confusion matrix for the full test set shows that the majority of misclassifications occurred among Cases A, B, and C, rather than between these three cases and either Case D or ``None reported.'' Of the 20 true Case A instances, 11 were correctly classified, with five misclassified as Case C and three as Case B. Of the 15 true Case B instances, eight were correctly classified, with six misclassified as Case C. Of the 23 true Case C instances, 17 were correctly classified, with three misclassified as Case A and two as Case B. Of the 11 true Case D instances, seven were correctly classified, with two misclassified as Case B and two as Case C. All 25 ``None reported'' instances were classified correctly, with zero instances of any illness case being misclassified as ``None reported'' or vice versa.

Feature importance analysis of the final Random Forest model identified \texttt{fell\_sick\_Yes} and \texttt{fell\_sick\_No} as the two most important features by a substantial margin, accounting for approximately 60\% of the total feature importance. This reflects the near-deterministic relationship between this field and the ``None reported'' class rather than a substantive predictor of illness type. Among the remaining features, importance was concentrated in the four disposal-method indicator variables, ranked in descending order: \texttt{disposal\_method\_Burn}, \texttt{disposal\_method\_Bush/Open Dump}, \texttt{disposal\_method\_Stream}, and \texttt{disposal\_method\_Bin}. Demographic features, including age, education level, occupation, sex, and sub-area of residence, contributed comparatively little to the model's predictions.

\subsection{Image Classifier Results}
\label{sec:image-results}

On the held-out test set ($N=415$), the fine-tuned MobileNetV2 model achieved a test accuracy of 88.2\% (test loss 0.349) and a macro F1 score of 0.87. The per-class performance was as follows: cardboard achieved precision 0.91, recall 0.95, F1 0.93 ($n=63$); glass achieved precision 0.91, recall 0.90, F1 0.91 ($n=80$); metal achieved precision 0.81, recall 0.96, F1 0.88 ($n=83$); paper achieved precision 0.92, recall 0.82, F1 0.87 ($n=89$); plastic achieved precision 0.92, recall 0.82, F1 0.87 ($n=74$); and trash achieved precision 0.77, recall 0.77, F1 0.77 ($n=26$), which was the weakest-performing class on all three metrics. The weighted-average F1 score across all classes was 0.88, closely tracking the macro F1 score, indicating that the performance was not driven disproportionately by any single high-support class.

The confusion matrix reveals two dominant, largely symmetric confusion pairs. Glass and metal were confused with each other in both directions: 5 of 80 true glass instances were misclassified as metal, and 2 of 83 true metal instances were misclassified as glass. Paper was confused with both cardboard and metal at comparable rates: of 89 true paper instances, 6 were misclassified as cardboard and 7 as metal, whereas cardboard and metal were rarely misclassified as paper (1 and 1 instance, respectively), indicating that this confusion was concentrated in one direction, with paper images resembling cardboard or metal more often than the reverse. Plastic showed a more diffuse error pattern, with its 13 misclassified instances (of 74 total) spread across glass (3), metal (4), paper (4), and trash (2), rather than concentrated in a single confused class. Trash, the smallest class in the dataset, had its 6 misclassified instances (of 26 total) split evenly between glass (2), metal (2), and plastic (2).

These error patterns are consistent with the visual similarity between material types: glass and metal objects can share reflective surface properties, and paper, cardboard, and certain metal packaging can share similar shape, color, and texture under varying lighting conditions rather than reflecting confusion between visually dissimilar categories. Trash, comprising a heterogeneous mixture of non-recyclable items rather than a single visually consistent material class, showed the most evenly distributed error pattern of any category, consistent with its lower precision and recall relative to the other five classes.

\section{Discussion}
\label{sec:discussion}
The results presented in Section~\ref{sec:results} bear on two related but distinct questions raised in this paper's introduction: whether the community-reported link between disposal method and illness identified in the original 2022 field investigation (Section~\ref{sec:study-area}) carries genuine, quantifiable predictive signal, and whether a simpler, camera-based alternative to the multi-sensor waste-sorting concept originally proposed can achieve practically useful accuracy. Taken together, the two models developed in this study offer a qualified yes to both questions, though with meaningfully different degrees of confidence. The survey predictive model demonstrates that disposal method carries real, measurable information about illness type once illness is present, moving the original study's case-based framework from a descriptive association to a testable, quantified relationship, though, as discussed below, this signal is more modest and more narrowly scoped than the original framework's headline association might suggest. The image classification model, by contrast, demonstrates a more straightforwardly strong result: an 88.2\% test accuracy and macro F1 of 0.87 (Section~\ref{sec:image-results}) indicate that a single trained model operating on ordinary camera images can perform waste-category sorting at a level of reliability that plausibly substitutes for the combination of infrared, capacitive, and inductive sensors envisioned in the original conceptual design. These two findings are complementary rather than symmetric: one provides cautious, partial support for a health-outcomes hypothesis under real-world data constraints, while the other provides a more confident technical validation of a specific engineering substitution.

This result requires careful interpretation, and it is worth stating plainly what the survey model shows and does not show. The dominant predictor in the final Random Forest model is not the disposal method but the \texttt{fell\_sick} field itself (Section~\ref{sec:survey-results}), which near-deterministically separates respondents who reported any illness from those who did not. This is an expected, largely tautological result rather than a substantive finding: knowing whether someone reported being sick is, unsurprisingly, highly predictive of whether they fall into the ``None reported'' category. The more meaningful test of this paper's central hypothesis lies not in the full 5-class result but in the sick-respondent subset (Section~\ref{sec:survey-results}), where the model must distinguish \emph{which} illness a respondent reported, given that they reported one at all. On this narrower and arguably more clinically relevant task, disposal method emerges as the leading substantive predictor, ahead of all demographic features, including age, education, occupation, sex, and sub-area of residence. The model's sick-subset performance (macro F1 = 0.63, accuracy = 0.62) indicates that disposal method carries a genuine, learnable signal about illness type, consistent with the case-based framework reported by residents in the original 2022 investigation. However, this performance is modest in absolute terms: correct classification of approximately three out of five is a meaningful improvement over chance for a four-class problem, but it falls well short of the reliability that would be required for disposal method alone to serve as a clinical or diagnostic predictor of illness type at the individual level. The confusion matrix (Section~\ref{sec:survey-results}) reinforces this modesty in a specific way: most misclassifications occurred among Cases A, B, and C rather than between these cases and Case D or ``None reported,'' suggesting that the model reliably distinguishes sick from healthy respondents but has a harder time distinguishing among the specific illness types associated with unsafe disposal practices, several of which, malaria from mosquito-breeding sites, cholera and dysentery from contaminated water, and headaches or nausea from smoke exposure, may plausibly co-occur or be difficult for respondents to self-report with precision in the first place.

The image classification results directly address a specific engineering question raised by the original 2022 study: whether the multi-sensor waste-sorting concept proposed, combining infrared detection, inductive metal sensing, capacitive material sensing, and optical glass detection across a multi-bin conveyor system (Concept 2, selected as the strongest of the three candidate designs in the original weighted decision matrix), is the most practical path to automated waste sorting in a resource-constrained community setting. The vision-based classifier developed in this study achieved 88.2\% test accuracy and macro F1 = 0.87 (Section~\ref{sec:image-results}), a level of performance comparable to mid-tier published results on the same benchmark dataset (Section~\ref{sec:related-work}), using a single camera and a trained model rather than four distinct physical sensing modalities. This substitution carries clear practical advantages: a camera-based system requires substantially less specialized hardware, avoids the calibration and maintenance demands of multiple sensor types operating in tandem, and can be deployed on commodity devices already present in many communities, including smartphones, rather than requiring a purpose-built sensing infrastructure. These advantages align with this paper's broader framing of automated sorting as a community-level rather than industrial-scale intervention, where the capital and maintenance costs associated with industrial sensor-based sorting systems (Section~\ref{sec:related-work}) are prohibitive.

This substitution is not without tradeoffs, which should be stated directly rather than implied. A vision-based classifier trained on TrashNet, a dataset of individually staged waste items photographed under comparatively controlled conditions, has not been validated on images of waste as it actually appears in Atonsu, which is often mixed, degraded by open burning or water exposure, or embedded within the specific disposal contexts documented in Section~\ref{sec:study-area}. In contrast, the physical multi-sensor design was conceived specifically around materials likely to appear in the local waste stream and does not depend on visual clarity or staged presentation. Whether the accuracy reported here would transfer to real Atonsu waste images, photographed under variable lighting and often partially obscured by dirt, moisture, or co-mingled materials, remains an open question that this study does not resolve; this limitation is discussed further in Section~\ref{sec:limitations}. Therefore, the vision-based approach is best understood not as a proven replacement for the original multi-sensor concept but as a technically validated and substantially lower-cost candidate for future field deployment and testing.

Both findings ultimately raise the same practical question: what would it take to move from a working model to an actual intervention in Atonsu or a comparable community? The citizen-science and community environmental monitoring literature reviewed in Section~\ref{sec:related-work} offers a cautionary benchmark here: among the reviewed programs that identified an environmental problem, only a minority followed through to reporting or corrective action \citep{castro2024cswater}, suggesting that data collection and model accuracy alone are insufficient without a clear institutional pathway to action. This concern is reinforced by the Ghana-specific governance literature reviewed in Section~\ref{sec:related-work}, which attributes persistent waste management failures in Kumasi primarily to budgetary and enforcement gaps rather than technical ones \citep{kumasiswm2019, kma_governance2021}. Read together, this suggests that neither model developed here is likely to improve outcomes in isolation: the survey model's value lies in giving municipal or community health planners a lightweight, data-grounded way to prioritize which disposal practices warrant the most urgent intervention, while the image classifier's value lies in lowering the technical barrier to deploying sorting infrastructure, not in resolving the institutional and funding constraints identified as the primary bottleneck in prior work.

\section{Limitations}
\label{sec:limitations}

Several limitations qualify the findings of this study, most of which are concentrated in the survey predictive model (Section~\ref{sec:survey-model}). First, the survey underlying this model was cross-sectional and captured disposal practices and reported illnesses at a single point in time. This design supports association but not causal inference; the model's ability to predict illness type from disposal method (Section~\ref{sec:survey-results}) demonstrates a learnable statistical relationship, not evidence that a given disposal practice causes a specific illness. Second, illness categories were self-reported by respondents rather than clinically or diagnostically confirmed, introducing the possibility of misclassification, recall bias, or conflating symptoms across the four case categories. This concern is reinforced by the confusion matrix pattern noted in Section~\ref{sec:discussion}, where misclassifications were concentrated among Cases A, B, and C rather than being randomly distributed. Third, the model's more clinically meaningful evaluation, distinguishing illness type among respondents who reported being sick, was conducted on a held-out subset of only 69 respondents (Section~\ref{sec:survey-results}), a sample size that limits the statistical precision of per-class estimates and constrains how confidently the reported macro F1 of 0.63 should be generalized beyond this specific sample. This limitation is compounded by the finding, discussed in Section~\ref{sec:discussion}, that the dominant feature in the full model is \texttt{fell\_sick} rather than the disposal method itself: the more substantively interesting task is effectively evaluated on a smaller and less statistically powered slice of the data than the headline dataset size ($N=470$) suggests. Finally, the disposal method may correlate with illness type partly through unmeasured or incompletely controlled confounding variables, including sub-area of residence, proximity to specific water sources, and socioeconomic factors that could independently influence both disposal behavior and health outcomes, which this model does not explicitly disentangle.

The image classification model has a distinct set of limitations, primarily centered on the domain gap between its training data and intended deployment context. The TrashNet dataset used to train and evaluate this model consisted of individually staged waste items photographed in isolation under relatively controlled lighting and background conditions. This differs substantially from the actual waste stream in Atonsu, which, as described in Section~\ref{sec:study-area}, is frequently mixed, degraded by open burning or prolonged water exposure, and embedded within informal disposal sites rather than being presented as discrete, clean objects. No on-site validation using images collected from Atonsu's actual waste stream has yet been conducted; the model's 88.2\% test accuracy and macro F1 of 0.87 (Section~\ref{sec:image-results}) should therefore be understood as an upper bound achieved under favorable, controlled conditions, with real-world deployment performance in Atonsu remaining an open empirical question rather than a demonstrated result. Relatedly, TrashNet's six material categories, cardboard, glass, metal, paper, plastic, and trash, were defined for a general recycling context and may not map precisely onto the specific waste composition or sorting priorities relevant to Atonsu, where the disposal-illness framework established in Section~\ref{sec:study-area} centers on organic and mixed household waste rather than primarily recyclable materials. Closing this domain gap through on-site data collection and model validation is identified as a priority direction for future research.

\section{Conclusion and Recommendations}
\label{sec:conclusion}
This study extended the community-reported disposal-illness framework established by a 2022 field investigation in Atonsu, Kumasi (Section~\ref{sec:study-area}) with two data-driven models. A Random Forest classifier demonstrated that disposal method is a genuine, learnable predictor of illness type among respondents who reported being sick, achieving a macro F1 score of 0.63 on a held-out subset of 69 respondents (Section~\ref{sec:results}), a modest but real result that moves the original study's case-based associations from description toward quantification, while stopping well short of individual-level diagnostic reliability. A MobileNetV2 image classifier, evaluated on the TrashNet benchmark, achieved 88.2\% test accuracy and a macro F1 score of 0.87 (Section~\ref{sec:results}), providing stronger technical evidence that vision-based waste sorting can substitute for the multi-sensor mechanical design originally proposed, at a substantially lower hardware cost and complexity (Section~\ref{sec:discussion}). Together, these results support the central claim of this study that community-level environmental and health challenges identified through direct fieldwork can be productively extended with modern data-driven methods without requiring a large-scale institutional data infrastructure to do so.

Building on the original 2022 investigation's recommendations, namely, citizen education on the health risks of improper waste disposal, increased government investment in waste removal infrastructure, and development of more advanced waste sorting technology, this study's findings suggest two specific refinements. First, given the survey model's finding that disposal method is the leading substantive predictor of illness type among sick respondents (Section~\ref{sec:results}), public health messaging in Atonsu and comparable communities would benefit from being disposal-method-specific rather than generic: residents disposing of waste through open burning, bush dumping, and stream dumping face distinguishable health risk profiles (Section~\ref{sec:study-area}), and communication campaigns tailored to each practice are likely to be more actionable than undifferentiated waste-safety messaging. Second, given the cost and complexity advantages of vision-based sorting established in Section~\ref{sec:discussion}, community waste management pilots in resource-constrained settings should prioritize camera-based classification approaches over multi-sensor mechanical designs as a first deployment step, reserving more complex sensor-based systems for contexts where budget and maintenance capacity are less constrained. However, neither recommendation is sufficient on its own: as discussed in Section~\ref{sec:discussion}, prior work on Ghana's municipal waste governance and citizen-science environmental monitoring both indicate that technical interventions require a concrete institutional pathway to funding, enforcement, or action to translate into improved outcomes.

Future work stemming from this study falls into three areas. First, on-site validation of the image classifier using photographs of waste as it actually appears in Atonsu, mixed, degraded by burning or water exposure, and embedded in real disposal contexts is needed to close the domain gap identified in Section~\ref{sec:limitations} and to determine whether the accuracy reported in Section~\ref{sec:results} transfers beyond the controlled conditions of the TrashNet benchmark. Second, longitudinal or panel survey data collection, following the same households or community members over time, would allow future work to move beyond the cross-sectional design's limits (Section~\ref{sec:limitations}) toward stronger evidence about whether disposal practices causally contribute to illness rather than merely correlating with it. Third, given that neither model developed here is sufficient in isolation (Section~\ref{sec:discussion}), a natural next step is to pilot both models together as a single community health and waste management tool, pairing disposal-method-based illness risk flagging with camera-based sorting support, rather than continuing to develop and evaluate them as separate outputs.

\section{Author Contributions}
\label{sec:author-contributions}

H.A.\ Osei is the lead and corresponding author of this study, having originated and directed the underlying 2022 field investigation and its extension into the present work.

\textbf{Hilda Adwubi Osei:} Conceptualization, Methodology, Software, Formal Analysis, Investigation, Data Curation, Writing -- Original Draft, Writing -- Review \& Editing, Visualization, Project Administration.

\textbf{Catherine Tenewaa Osei:} Conceptualization (health outcomes framing), Investigation, Writing -- Review \& Editing.

\textbf{Desdemona Yaa Asobayire:} Investigation (mechanical and sensor-based waste management systems, Related Work), Writing -- Review \& Editing.

All authors read and approved the final manuscript.

\bibliographystyle{unsrtnat}
\bibliography{references}

\appendix
\section{Survey Questionnaire}
\label{app:questionnaire}
The following is the original questionnaire instrument administered during the 2022 field investigation described in Section~\ref{sec:study-area}, reproduced here in full for reference and reproducibility.

\bigskip
\noindent\textit{This questionnaire is meant to obtain information from the residents in Atonsu Bukuro. It seeks to investigate matters arising from improper waste management in Atonsu Bukuro. Please tick where necessary.}

\bigskip
\begin{enumerate}
  \item Sex
    \begin{itemize}
      \item Male
      \item Female
    \end{itemize}

  \item What is your age?

  \item What is your educational background?
    \begin{itemize}
      \item Junior high school
      \item Senior high school
      \item Tertiary
      \item Illiterate
    \end{itemize}

  \item Are you a resident at Atonsu?
    \begin{itemize}
      \item Yes
      \item No
    \end{itemize}

  \item What's your occupation?
    \begin{itemize}
      \item Student
      \item Entrepreneur
      \item Civil servant
      \item Employed
      \item Unemployed
    \end{itemize}

  \item Do you think improper waste management is a problem at Atonsu?

  \item If yes, please state your reason.

  \item How do you dispose your waste?

  \item Do you think your method of disposal is harmful?

  \item If yes, has it affected you?

  \item Do you fall sick?
    \begin{itemize}
      \item Yes
      \item No
    \end{itemize}

  \item If yes, what do you often complain of?

  \item How do you want the improper waste management in Atonsu Bukuro to be solved?

  \item Would you readily prefer a method of control that does not pose a threat to the environment and residents?
    \begin{itemize}
      \item Yes
      \item No
    \end{itemize}

  \item Please give a comment you think will be useful to the project.
\end{enumerate}

\section{Additional Figures}
\label{app:figures}

This appendix presents supporting figures generated during model development and referenced in Section~\ref{sec:results} but not fully reproduced in the main text.

\begin{figure}[H]
  \centering
  \includegraphics[width=0.8\linewidth]{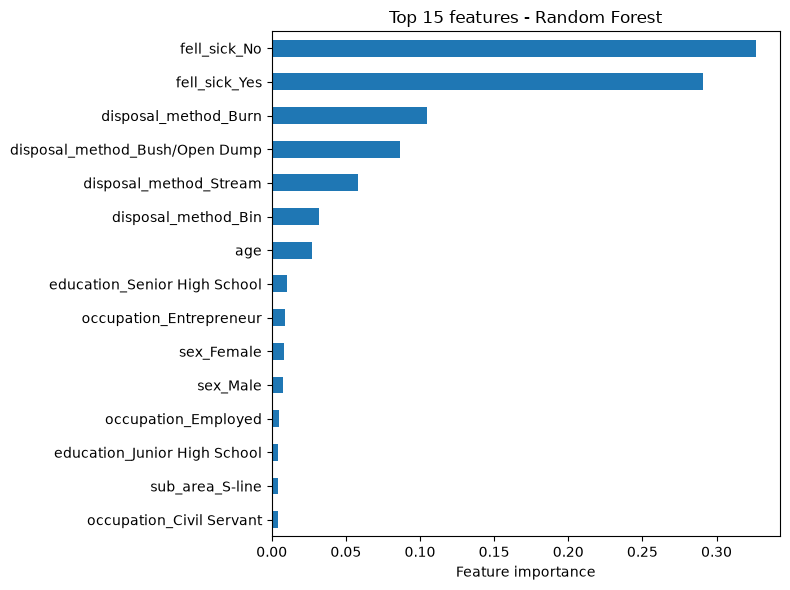}
  \caption{Top 15 feature importances for the final Random Forest survey model, showing \texttt{fell\_sick\_Yes} and \texttt{fell\_sick\_No} dominating, followed by the four disposal-method indicator features in descending order (see Section~\ref{sec:survey-results}).}
  \label{fig:feature-importance}
\end{figure}

\begin{figure}[H]
  \centering
  \includegraphics[width=0.8\linewidth]{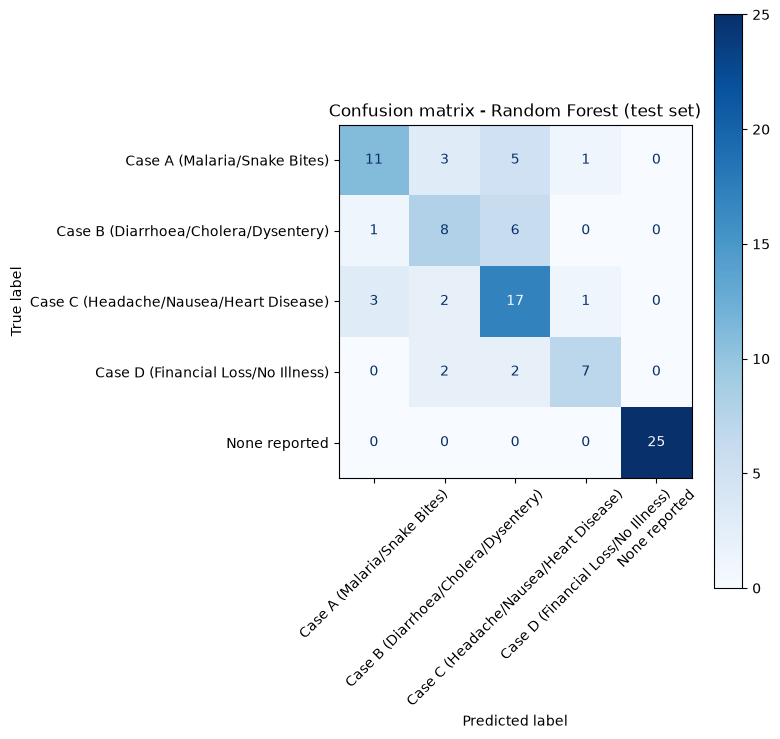}
  \caption{Confusion matrix for the survey predictive model on the full held-out test set ($N=94$), covering all five categories including ``None reported'' (see Section~\ref{sec:survey-results}).}
  \label{fig:survey-cm}
\end{figure}

\begin{figure}[H]
  \centering
  \includegraphics[width=0.8\linewidth]{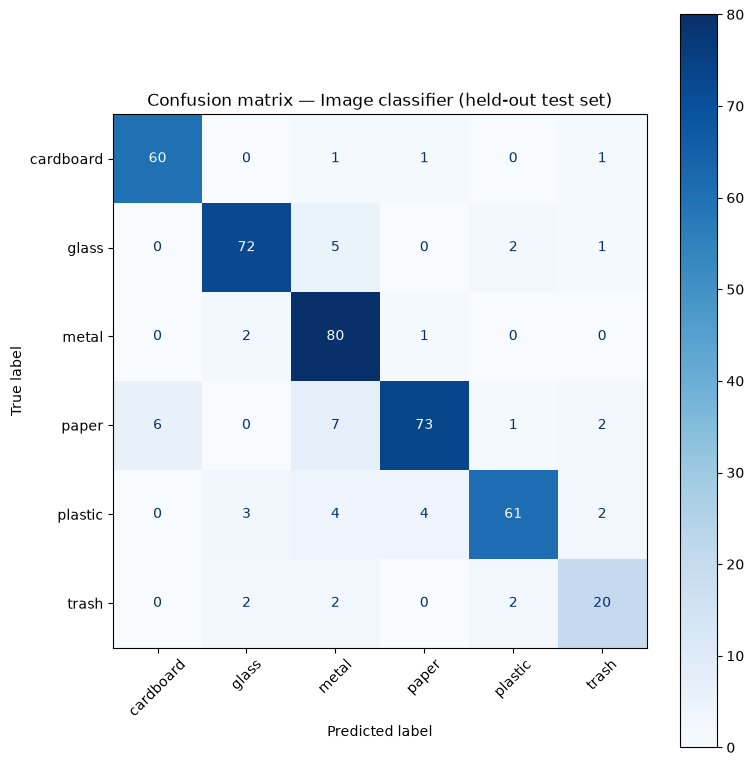}
  \caption{Confusion matrix for the image classification model on the held-out test set ($N=415$), covering all six TrashNet waste categories (see Section~\ref{sec:image-results}).}
  \label{fig:image-cm}
\end{figure}

\section{Data and Code Availability}
\label{app:data}
All code used in this study, including the survey predictive model and image classification model notebooks, along with generated figures, is publicly available at \url{https://github.com/hildaadwubiosei/ATONSU-WASTE-PROJECT}. Code is released under the MIT License; figures and results are released under a CC BY 4.0 license, as detailed in the repository's README. The TrashNet dataset used to train and evaluate the image classification model is publicly available and cited in Section~\ref{sec:methodology}. The original survey data collected during the 2022 field investigation (Section~\ref{sec:study-area}) is not shared publicly in raw form, as respondent-level household location data was collected alongside survey responses and its public release could compromise respondent privacy even in the absence of directly identifying information such as names; a de-identified, aggregated version of the dataset may be made available upon reasonable request to the corresponding author.

\end{document}